\pdfoutput=1
\RequirePackage{float}
\documentclass{bmvc2k}
\usepackage{amsmath}
\usepackage{amssymb}
\usepackage{booktabs}
\usepackage{multirow}
\usepackage{graphicx}
\usepackage{cleveref}
\usepackage{wrapfig}
\usepackage{algorithm}
\usepackage[expansion=false]{microtype}
\crefname{figure}{Figure}{Figures}\Crefname{figure}{Figure}{Figures}
\crefname{table}{Table}{Tables}\Crefname{table}{Table}{Tables}
\crefname{equation}{Eq.}{Eqs.}\Crefname{equation}{Eq.}{Eqs.}
\crefname{section}{Section}{Sections}\Crefname{section}{Section}{Sections}

\title{EyeTAG: Eye Trajectory-Aware Gaze Estimation}

\addauthor{Jungmin Lee}{jungmin@soongsil.ac.kr}{1}
\addauthor{Niamat Ullah}{niamat016@ssu.ac.kr}{2\textdagger}
\addauthor{Yoseob Han}{yoseob.han@ssu.ac.kr}{2\textdagger}
\addinstitution{
Department of Information and Telecommunication Engineering\\
Soongsil University\\
Seoul, Republic of Korea
}
\addinstitution{
Department of Electronic Engineering\\
Soongsil University\\
Seoul, Republic of Korea
}

\runninghead{Lee et al.}{EyeTAG: Eye Trajectory-Aware Gaze Estimation}

\def\etal{\emph{et al}\bmvaOneDot}

\begin{document}
\maketitle

\begingroup
\renewcommand{\thefootnote}{\textdagger}
\footnotetext{Corresponding authors.}
\endgroup
\begin{abstract}
Gaze estimation under natural head-eye motion underpins applications from driver
monitoring to human-computer interaction. Single-frame methods predict each frame
independently, so consecutive outputs fluctuate as jitter. Multi-frame methods
reduce this, but they learn motion \emph{implicitly} inside appearance features,
so the gaze trajectory is never an explicit variable. We propose \textbf{EyeTAG}
(\textbf{Eye} \textbf{T}rajectory-\textbf{A}ware \textbf{G}aze Estimation), a
causal multi-frame framework built around an \emph{explicit} first-order gaze
prior: at each step it differentiates its own recent predictions and feeds the
resulting trajectory back as a compact kinematic token. Because differencing is
translation-invariant in gaze space, this token carries subject-invariant motion
rather than personal gaze offsets. Face and eye streams supply visual evidence,
fused by cross-attention and a causal Transformer decoder. EyeTAG reduces the mean
angular error by about 1.0\textdegree~on Gaze360 and performs on par with the
strongest baseline on EVE (2.56\textdegree~vs.\ 2.58\textdegree).
Within-model ablations, which keep the encoder and the rest of the architecture fixed and vary only the gaze history, show that the differential formulation, rather than temporal context alone, removes the systematic saccade bias that persists even with an absolute gaze-history prior. Our code is available at \url{https://github.com/peter8366/EyeTAG}.
\end{abstract}

\section{Introduction}
\label{sec:intro}
Gaze estimation is the task of predicting where a person is looking, and it is fundamental for understanding human attention and for enabling natural human-computer interaction. It is used in many applications, such as driver monitoring \cite{sharma2024review, shah2022driver}, behavioral and cognitive analysis \cite{sunny2021eye}, and gaze-driven interfaces for extended reality \cite{pfeuffer2017gaze+, lee2022vrdoc}. Despite steady progress, achieving robust performance in unconstrained real-world conditions is still challenging, because the head pose, eye appearance, occlusion, and illumination vary widely.

Early gaze-estimation methods relied on \textbf{geometric modeling} of the eyeball and corneal reflections. These methods are accurate in controlled environments, but they require user-specific calibration and are sensitive to occlusion and illumination. In contrast, \textbf{appearance-based approaches} use deep neural networks to learn a direct mapping from facial and ocular appearance to 3D gaze. Large-scale in-the-wild datasets such as GazeCapture \cite{krafka2016eye} and Gaze360 \cite{kellnhofer2019gaze360} have allowed these methods to generalize across diverse subjects and scenes.

\begin{figure*}[t]
  \centering
  \includegraphics[width=1.0\linewidth]{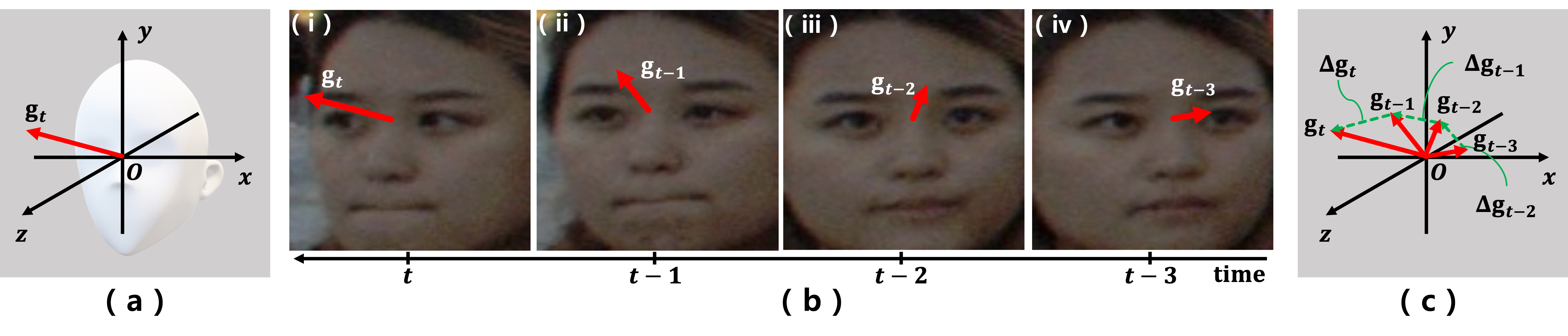}
  \caption{Visualization of gaze direction $\mathbf{g}_t$ and its underlying motion kinematics. (a) Definition of the 3D coordinate system. (b) Four consecutive frames illustrating the gaze vector evolving over time. (c) Overlaid 3D visualization, where $\Delta \mathbf{g}_t$ denotes the differential change (i.e., velocity) between consecutive gaze directions.}
  \label{fig:coord}
\end{figure*}

Most appearance-based models predict gaze from a single image, so consecutive predictions are statistically independent and fluctuate as unrealistic jitter or physically implausible outliers during rapid head-eye movements. Temporal continuity itself is not a new concern: multi-frame methods have addressed it with recurrent layers~\cite{kellnhofer2019gaze360}, temporal attention~\cite{hu2025fifa}, and sequence models over per-frame features~\cite{jindal2024spatio}. However, these methods represent motion only \emph{implicitly}, inside high-dimensional appearance features --- the gaze trajectory is never carried as an explicit variable, and the model must rediscover it at every step. Human gaze, in contrast, changes smoothly over time \cite{lisberger2015visual} and exhibits temporally coherent dynamics. As illustrated in \cref{fig:coord}, we model the local first-order change of the 3D gaze vector $\mathbf{g}_t \in \mathbb{R}^3$ as:
\begin{equation}
\mathbf{g}_t = \mathbf{g}_{t-1} + \Delta \mathbf{g}_t,
\label{eq:gaze_dynamics}
\end{equation}
where $\Delta \mathbf{g}_t$ denotes the differential change between two consecutive frames. This temporal continuity means that, for stable estimation, it is important to condition the current prediction on the recent gaze kinematics, that is, on the sequence of past differentials rather than on appearance alone.

To this end, we propose \textbf{EyeTAG} (\textbf{Eye} \textbf{T}rajectory-\textbf{A}ware \textbf{G}aze Estimation), a causal multi-frame framework whose defining component is an explicit, autoregressive gaze prior. Unlike previous temporal methods, which learn motion implicitly from high-dimensional appearance features, EyeTAG takes its own recent predictions, converts them into first-order differences, and feeds this differential trajectory back into the model as \textbf{(i) a kinematic prior} encoded by a Multi-Layer Perceptron (MLP). The differencing operator is translation-invariant in gaze space, so a constant offset applied to the whole trajectory leaves the prior unchanged; the prior therefore carries the motion patterns that are shared across subjects and discards subject-specific resting-gaze offsets. To ground this prior in reliable visual evidence, EyeTAG additionally encodes \textbf{(ii) global facial context} with a ResNet-50 and \textbf{(iii) fine-grained ocular cues} with a ResNet-18, and a fusion module based on cross-attention and a causal Transformer decoder integrates the three streams. Both the visual window and the prediction history contain only past and current information, so the whole model is causal and can be run in a streaming setting.

Our main contributions are summarized as follows:
\begin{itemize}
    \item We introduce an \emph{explicit} first-order gaze prior for video gaze estimation. Rather than delegating temporal structure to a visual sequence model, EyeTAG differentiates its own recent predictions and feeds the resulting trajectory back as a dedicated, subject-invariant motion token.
    \item We design a causal fusion architecture around this prior, which aligns global facial context with fine-grained ocular cues through unidirectional cross-attention and aggregates them with the kinematic token under a causal Transformer decoder, so that inference uses only past and current frames.
    \item Beyond average angular error, we analyze temporal behavior explicitly. Within-model ablations that keep the encoder and the rest of the architecture fixed identify the removal of systematic saccade bias as the specific effect of the differential prior, rather than a by-product of temporal smoothing.
\end{itemize}

\section{Related Works}
\label{sec:related}

\subsection{Gaze Estimation}
\medskip\noindent
\textbf{Geometric-based Methods.} Early gaze-estimation approaches modeled the 3D geometry of the eyeball and corneal reflections \cite{hansen2009eye, guestrin2006general}. These methods can achieve high accuracy in controlled environments, but they usually require user-specific calibration and are sensitive to illumination and occlusion. They provide explicit physical constraints for gaze estimation, but because they depend on specialized hardware and constrained acquisition conditions, they are difficult to apply in real-world scenarios.

\noindent\textbf{Appearance-based Methods.} With the development of deep learning, appearance-based methods have become the preferred approach \cite{krafka2016eye, cheng2018appearance, kellnhofer2019gaze360, cheng2022gaze, abdelrahman2023l2cs, xiao20252gaze}. These methods learn a direct mapping from visual appearance to the 3D gaze direction without geometric calibration. In addition, large-scale in-the-wild datasets such as GazeCapture \cite{krafka2016eye} and Gaze360 \cite{kellnhofer2019gaze360} have allowed these data-driven methods to generalize across diverse subjects and scenes. Cheng \etal~\cite{cheng2018appearance} proposed an asymmetric regression that uses facial appearance and improves accuracy in single-frame settings. More recently, Abdelrahman \etal~\cite{abdelrahman2023l2cs} used a multi-branch architecture that regresses the pitch and yaw angles separately for more fine-grained estimation. Cheng and Lu~\cite{cheng2022gaze} proposed GazeTR, which uses a hybrid CNN-Transformer architecture to capture long-range spatial dependencies from facial images. Xiao \etal~\cite{xiao20252gaze} further improved appearance-based estimation using deformable sparse attention, but this method still does not perform explicit temporal reasoning.

\subsection{Temporal and Multi-frame Modeling}

Temporal modeling for gaze estimation has been widely explored. Kellnhofer \etal~\cite{kellnhofer2019gaze360} used an LSTM to capture temporal continuity, Hu \etal~\cite{hu2025fifa} introduced temporal attention over informative frames, and Jindal \etal~\cite{jindal2024spatio} proposed STAGE to combine inter-frame motion cues with global video dynamics. Other works model head-face-eye interactions~\cite{guan2023end} or eye-head-body coordination~\cite{nonaka2022dynamic}, while related approaches forecast future fixations from scene context~\cite{hu2020dgaze, hu2021fixationnet}.

Unlike these methods, where temporal dynamics are represented implicitly through appearance features or contextual cues, EyeTAG explicitly models gaze motion itself. Our Trajectory-Aware Kinematic Encoder (TAKE) differentiates past gaze predictions and feeds the resulting compact kinematic prior back autoregressively, providing explicit differential feedback for temporally stable gaze estimation.

\subsection{Cross-modal Fusion of Face and Eye Features}
Accurate gaze estimation requires an effective combination of global facial context and local ocular motion. The facial region provides coarse cues related to head pose, while the eye region captures fine-grained pupil and iris motion. Early multi-branch architectures combined these cues by simple feature concatenation, but this approach often failed to align the two modalities when they were misregistered.

Fischer \etal~\cite{fischer2018rt} extracted features from the left and right eyes using separate VGG-based branches and combined them with facial representations for gaze prediction. Cheng \etal~\cite{cheng2020coarse} later proposed a coarse-to-fine adaptive network that first estimates a coarse gaze direction from the face and then refines it using a residual computed from the eye image. More recent cross-modal interaction frameworks, such as CrossGaze \cite{catruna2024crossgaze}, use cross-attention to fuse face and eye features.

In short, prior work either relies on appearance cues alone, which can lead to temporal jitter, or models temporal dynamics implicitly through high-dimensional appearance features. EyeTAG instead explicitly models gaze kinematics and fuses the resulting prior with multi-scale face and eye features.

\begin{figure*}[t]
    \centering
    \includegraphics[width=1.0\textwidth]{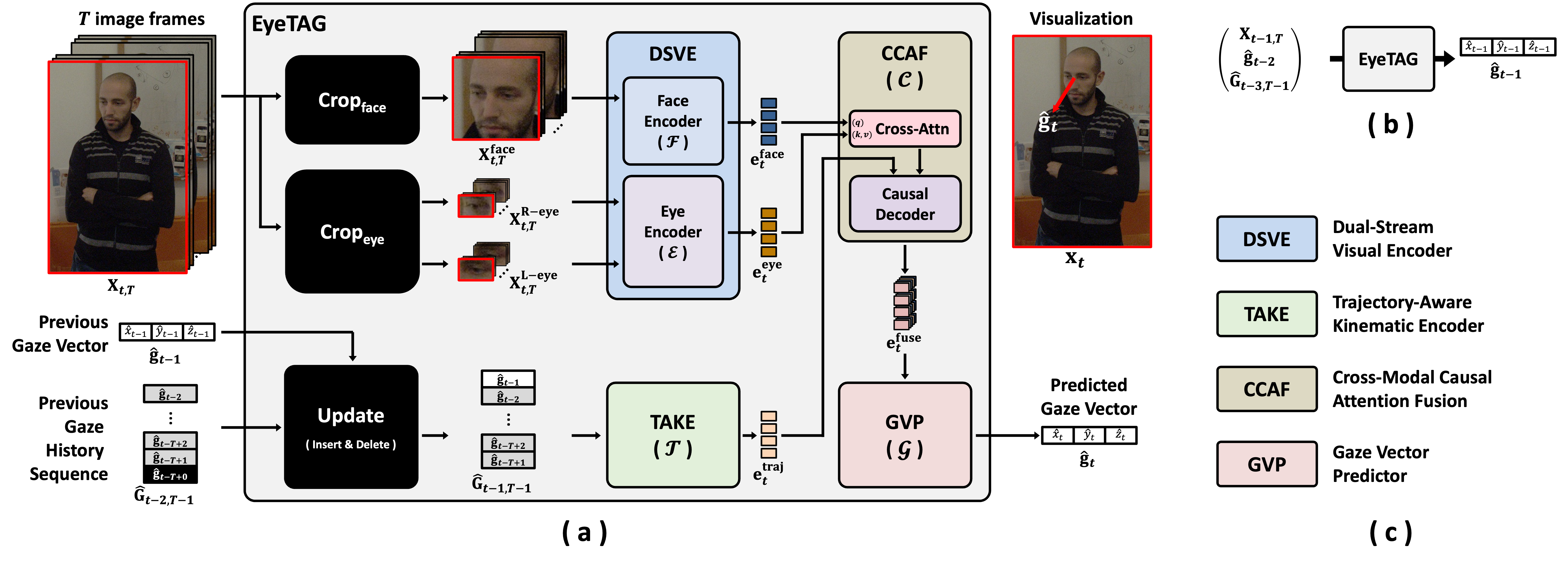}
    \caption{Overview of the EyeTAG framework and its autoregressive components.
    (a) Main architecture. The DSVE ($\mathcal{F}, \mathcal{E}$) extracts face and eye embeddings from multi-frame inputs, while the TAKE ($\mathcal{T}$) encodes temporal gaze-motion priors from previous predictions. These features are fused by the Cross-Modal Causal Attention Fusion (CCAF; $\mathcal{C}$) module with a causal Transformer decoder, followed by the Gaze Vector Predictor (GVP; $\mathcal{G}$) to estimate the final 3D gaze vector $\hat{\mathbf{g}}_t$. (b) Autoregressive update loop. (c) Legend for the modules used in the diagram.
    }
    \label{fig:method_overview}
\end{figure*}

\section{Method}
\label{sec:method}
Let $\mathbf{X}_{t,T} = \{\mathbf{x}_{t-T+1},~\cdots, \mathbf{x}_{t} \}$ denote a sequence of $T$ consecutive frames up to and including the current frame $\mathbf{x}_{t}\in\mathbb{R}^{3 \times H \times W}$. The goal of EyeTAG is to estimate the 3D gaze direction $\hat{\mathbf{g}}_t \in \mathcal{S}^2$ of the current frame $\mathbf{x}_t$. To do this, it uses both the spatial appearance and the temporal kinematic motion in the recent video context, while keeping the autoregressive consistency between consecutive predictions.

As shown in \cref{fig:method_overview}, EyeTAG consists of three core components and a lightweight prediction head:
\textbf{(i) Dual-Stream Visual Encoder (DSVE)}, which extracts per-frame facial and ocular representations using a VGGFace2-pretrained ResNet-50 for the face stream and a shared ImageNet-pretrained ResNet-18 for the eye streams, and stacks them along the temporal axis into token sequences;
\textbf{(ii) Trajectory-Aware Kinematic Encoder (TAKE)}, which models gaze dynamics by encoding differential gaze trajectories $\Delta \hat{\mathbf{G}}$ into a compact motion token through a lightweight MLP; and
\textbf{(iii) Cross-Modal Causal Attention Fusion (CCAF)}, which integrates the facial, ocular, and kinematic representations using cross-modal attention and a causal Transformer decoder. A lightweight Gaze Vector Predictor (GVP) then regresses the final 3D gaze vector $\hat{\mathbf{g}}_t$ from this fused representation.

\subsection{Dual-Stream Visual Encoder (DSVE)}
\label{subsec:stve}
The DSVE extracts both global facial context and local ocular appearance from consecutive video frames. Given an input sequence $\mathbf{X}_{t,T}\in\mathbb{R}^{T \times 3 \times H \times W}$ , we crop two spatially aligned sub-sequences to separate the facial region and the eye regions:
\begin{align}
\mathbf{X}_{t,T}^{\mathrm{face}} &= \mathrm{Crop}_{\mathrm{face}}\left( \mathbf{X}_{t,T} \right), \nonumber \\
(\mathbf{X}_{t,T}^{\mathrm{L-eye}},~\mathbf{X}_{t,T}^{\mathrm{R-eye}})
&= \mathrm{Resize}\left(\mathrm{Crop}_{\mathrm{eye}}\left( \mathbf{X}_{t,T} \right)\right).
\end{align}
Here, $\mathbf{X}_{t,T}^{\mathrm{face}}\in\mathbb{R}^{T \times 3 \times H_{\mathrm{face}} \times W_{\mathrm{face}}}$ is the full facial $T$ sequence of RGB frames, which captures the global head motion, and $(\mathbf{X}_{t,T}^{\mathrm{L-eye}}, \mathbf{X}_{t,T}^{\mathrm{R-eye}}) \in \mathbb{R}^{T \times 3 \times H_{\mathrm{eye}} \times W_{\mathrm{eye}}}$ are the localized left and right eye sequences. The eye sequences are cropped from the original frames and resized to the same spatial resolution as the facial input. These streams are processed by separate ResNet-based branches to extract region-specific appearance features, and the long-range temporal dependencies are modeled later by the CCAF module.

\begin{figure}[!tb]
    \centering
    \includegraphics[width=1.0\linewidth]{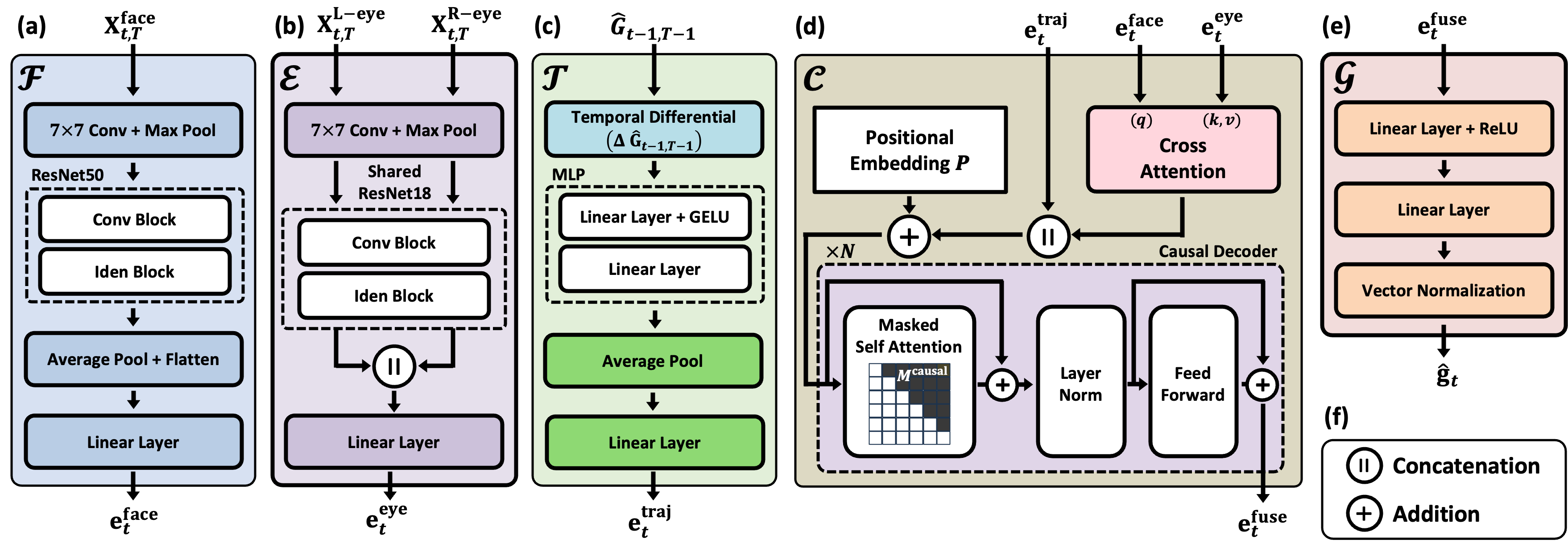}
    \caption{Detailed architecture of the EyeTAG framework. (a),(b) DSVE ($\mathcal{F}$, $\mathcal{E}$) and (c) TAKE ($\mathcal{T}$) modules generate visual features (global face, local eyes) and a kinematic prior (from $\Delta\hat{\mathbf{G}}$), respectively.
    These are fused by (d) CCAF ($\mathcal{C}$) and regressed by (e) GVP ($\mathcal{G}$) to produce the final gaze $\hat{\mathbf{g}}_t$. (f) Notation for operations used in the diagram.
    }
    \label{fig:architecture}
\end{figure}

\medskip\noindent
\textbf{Face Encoder.}
The face encoder $\mathcal{F}$ uses a ResNet-50~\cite{he2016deep} pretrained on VGGFace2~\cite{cao2018vggface2}, as shown in \cref{fig:architecture}(a). This face-domain pretraining is important because Gaze360 is captured under unconstrained real-world conditions, and ImageNet-pretrained features often do not generalize well to extreme head poses and diverse identities (we provide a detailed ablation of this design choice in Sec.~S2 of the supplementary material). Each frame is processed independently with shared weights, and the resulting features are stacked along the temporal axis to form a per-frame token sequence. The final linear layer of $\mathcal{F}$ maps these features to the shared latent dimension $D$:
\begin{align}
\mathbf{e}^\mathrm{face}_t
=
\mathcal{F}\!\left( \mathbf{X}_{t,T}^\mathrm{face} \right),
~~~\mathbf{e}^\mathrm{face}_t \in \mathbb{R}^{T \times D}.
\end{align}
Note that this per-frame design does not perform temporal modeling inside the DSVE. Instead, the temporal aggregation is done later in the CCAF. Because of this separation, the face encoder can focus on spatial representation learning, while the temporal dependencies are modeled by the cross-attention and the causal decoder.

\medskip\noindent
\textbf{Eye Encoder.}
The eye encoder $\mathcal{E}$ in \cref{fig:architecture}(b) is complementary to the face branch, and it focuses on fine-grained ocular appearance such as the iris position and the eyelid configuration. We use a lightweight ResNet-18~\cite{he2016deep} pretrained on ImageNet, and its weights are \emph{shared} between the left and right eye streams. This design reflects the natural symmetry of ocular appearance and reduces the number of parameters in the eye branch (we compare alternative eye-encoder backbones in Sec.~S3 of the supplementary material). Each eye is encoded independently, and the two per-eye descriptors are concatenated along the channel dimension at every time step. The final linear layer of $\mathcal{E}$ then maps the concatenated ocular features to the shared latent dimension $D$ used in the CCAF:
\begin{align}
\mathbf{e}^\mathrm{eye}_t = \mathcal{E} \left(
\mathbf{X}_{t,T}^\mathrm{L-eye},~
\mathbf{X}_{t,T}^\mathrm{R-eye}
\right),
~~~\mathbf{e}^\mathrm{eye}_t \in \mathbb{R}^{T \times D}.
\end{align}
As in the face branch, no temporal aggregation is applied at this stage.

Together, the facial and ocular encoders form the DSVE. They produce multi-scale per-frame visual feature representations, which are used as the visual input for the following kinematics-aware fusion module.

\subsection{Trajectory-Aware Kinematic Encoder (TAKE)}
\label{subsec:take}

The DSVE captures appearance from the face and eye sequences ($\mathbf{X}_{t,T}^\mathrm{face}$, $\mathbf{X}_{t,T}^\mathrm{eye}$), but it does not explicitly model how the gaze direction changes over time. To make this motion explicit, we introduce the TAKE module, shown in \cref{fig:architecture}(c). It encodes the differential gaze trajectories $\Delta \hat{\mathbf{G}}$ into a compact temporal prior that represents the continuity of eye motion.

At time step $t$, we keep a sliding window of the most recent $T-1$ predicted gaze vectors:
\begin{align}
\hat{\mathbf{G}}_{t-1, T-1} = [\hat{\mathbf{g}}_{t-1}, \hat{\mathbf{g}}_{t-2}, \cdots, \hat{\mathbf{g}}_{t-T+1}], ~~~\hat{\mathbf{G}}_{t-1, T-1} \in \mathbb{R}^{(T-1) \times 3},
\end{align}
where the oldest element is removed and the newest prediction is added as the window advances.

Instead of using raw gaze vectors, TAKE uses differential representations, which encode the relative motion between consecutive frames:
\begin{align}
\Delta \hat{\mathbf{g}}_{t-k} = \hat{\mathbf{g}}_{t-k} - \hat{\mathbf{g}}_{t-k-1},~~~k=1, \cdots, T-2.
\end{align}

This differential formulation is effective because the differencing operator removes subject-specific gaze offsets and keeps the relative motion between consecutive frames. As a result, the differential representation forms a compact, zero-centered distribution that preserves shared motion dynamics and improves the autoregressive stability as the context length $T$ increases, which we analyze quantitatively in \cref{subsec:ablation_prevgaze}.

We denote the resulting differential sequence as
\begin{align}
\Delta \hat{\mathbf{G}}_{t-1, T-1} = [\Delta \hat{\mathbf{g}}_{t-1}, \cdots, \Delta \hat{\mathbf{g}}_{t-T+2}], ~~~\Delta \hat{\mathbf{G}}_{t-1, T-1} \in \mathbb{R}^{(T-2) \times 3}.
\end{align}
The compact two-layer MLP $\mathcal{T}$ first flattens the delta tokens and then maps them into a single kinematic embedding at the shared latent dimension $D$ used in the CCAF:
\begin{align}
\mathbf{e}^\mathrm{traj}_t
=
\mathcal{T}\!\left( \Delta \hat{\mathbf{G}}_{t-1, T-1} \right),
~~~\mathbf{e}^\mathrm{traj}_t \in \mathbb{R}^{1 \times D}.
\end{align}

The resulting kinematic embedding $\mathbf{e}^\mathrm{traj}_t$ is then fused with the visual representations ($\mathbf{e}^\mathrm{face}_t,~\mathbf{e}^\mathrm{eye}_t$) in the CCAF module for kinematics-aware gaze prediction. Because it is built from differential gaze evolution instead of absolute gaze coordinates, it gives EyeTAG a compact and autoregression-friendly prior over gaze velocity and smoothness.

\subsection{Cross-Modal Causal Attention Fusion (CCAF)}
\label{subsec:cmaf}
As shown in \cref{fig:architecture}(d), the CCAF module integrates the per-frame visual features $(\mathbf{e}_{t}^\mathrm{face},\mathbf{e}_{t}^\mathrm{eye})$ from the DSVE and the kinematic prior $\mathbf{e}_{t}^\mathrm{traj}$ from the TAKE, in order to jointly reason about the current gaze direction. The visual features encode per-frame appearance cues, while the kinematic embedding provides the motion context of how the gaze changes over time. The CCAF works in two stages: a \emph{cross-modal} stage that aligns the facial and ocular streams using attention, and then a \emph{causal temporal} stage that aggregates the aligned visual sequence together with the kinematic prior.

\medskip\noindent
\textbf{Cross-modal Attention.}
In the first stage, the facial stream queries the ocular stream through a multi-head cross-attention operation. This allows the model to enrich the global head-pose representation with the fine-grained ocular evidence that is most relevant at each frame:
\begin{equation}
\begin{gathered}
\mathbf{e}_{t}^\mathrm{face \rightarrow eye}
=\mathrm{Att}_\mathrm{cross}(Q=\mathbf{e}_{t}^\mathrm{face},~K=V=\mathbf{e}_{t}^\mathrm{eye}), \\
\mathbf{v}_{t}=\mathrm{LayerNorm}\!\left(\mathbf{e}_{t}^\mathrm{face} + \mathbf{e}_{t}^\mathrm{face \rightarrow eye}\right),
~~~\mathbf{v}_{t} \in \mathbb{R}^{T \times D}.
\end{gathered}
\label{eq:cross_att}
\end{equation}
The residual connection and the layer normalization preserve the facial query signal, while the ocular evidence selectively re-weights the per-frame face features. This produces an aligned per-frame visual token sequence $\mathbf{v}_t$, which already encodes the head-eye coordination.

\medskip\noindent
\textbf{Causal Temporal Aggregation.}
In the second stage, we concatenate the aligned visual tokens $\mathbf{v}_t$ with the kinematic-prior token $\mathbf{e}_{t}^\mathrm{traj}$, add a learnable positional embedding $\mathbf{P}\in\mathbb{R}^{(T+1)\times D}$, and process the resulting sequence with a causal Transformer decoder~\cite{vaswani2017attention}:
\begin{equation}
\begin{gathered}
\mathbf{Z}_{t} =
[\mathbf{e}_{t}^\mathrm{traj} \parallel \mathbf{v}_{t}] + \mathbf{P}, \\
\mathbf{e}_{t}^{\mathrm{fuse}} =
\mathrm{CausalDecoder}\!\left(\mathbf{Z}_{t};~\mathbf{M}_\mathrm{causal}\right),
~~~\mathbf{e}_{t}^{\mathrm{fuse}} \in \mathbb{R}^{(T+1)\times D}.
\end{gathered}
\label{eq:causal}
\end{equation}
The causal mask ensures that the representation of frame $t$ depends only on the past observations and the kinematic prior, so that no future information leaks during autoregressive inference.

\medskip\noindent
\textbf{Gaze Regression.}
The final representation of the current frame is taken from the last token of the decoder output $\mathbf{e}_{t}^{\mathrm{fuse}}$, which we denote as $\mathbf{h}_{t} = \mathbf{e}_{t}^{\mathrm{fuse}}[-1] \in \mathbb{R}^{D}$. A lightweight regression head $\mathcal{G}$ then predicts the final 3D gaze direction:
\begin{align}
\mathbf{\hat{g}}_{t} &= \mathcal{G} \left( \mathbf{h}_{t} \right).
\end{align}

Through this two-stage fusion process, the CCAF allows EyeTAG to jointly reason about the visual appearance ($\mathbf{e}_{t}^\mathrm{face},\mathbf{e}_{t}^\mathrm{eye}$) and the gaze kinematics $\mathbf{e}^\mathrm{traj}_t$ under a strict causal constraint, so that the gaze predictions are not only spatially precise but also temporally coherent, autoregression-stable, and smooth across frames.
\section{Experiments}
\label{sec:experiment}

\begin{table}[t]
\centering
\footnotesize
\setlength{\tabcolsep}{6pt}
\begin{tabular}{l@{\hspace{1.5em}}ccc@{\hspace{1.5em}}ccc}
\toprule
\multirow{2}{*}{\textbf{Split}} & \multicolumn{3}{c}{\textbf{(a) Gaze360 Dataset~\cite{kellnhofer2019gaze360}}} & \multicolumn{3}{c}{\textbf{(b) EVE Dataset~\cite{park2020towards}}} \\
\cmidrule(lr){2-4}\cmidrule(lr){5-7}
 & \textbf{\# Clips} & \textbf{\# Frames} & \textbf{Avg Length (s)} & \textbf{\# Clips} & \textbf{\# Frames} & \textbf{Avg Length (s)} \\
\midrule
Train & 321 & 84,902 & 8.8 & 3,015 & 1,286,850 & 13.2 \\
Validation & 85 & 11,318 & 4.4 & 336 & 155,640 & 13.8 \\
Test & 68 & 16,031 & 7.9 & 387 & 169,860 & 14.0 \\
\bottomrule
\end{tabular}
\caption{Summary statistics of the (a) Gaze360 and (b) EVE datasets used in our experiments. For each split we report the number of clips (\# Clips), the number of annotated frames (\# Frames), and the average clip length in seconds (Avg Length (s)).}
\label{tab:dataset_stats}
\end{table}

\subsection{Dataset}
\label{subsec:dataset}

We conduct experiments on the Gaze360~\cite{kellnhofer2019gaze360} and EVE~\cite{park2020towards} datasets.
Gaze360 was captured from 238 participants under natural conditions in seven environments at 30 Hz, and it covers a wide head-pose range of $\pm 180^\circ$ yaw and $\pm 90^\circ$ pitch. Following the official split protocol~\cite{kellnhofer2019gaze360}, we use 84{,}902 training, 11{,}318 validation, and 16{,}031 test annotated frames, which are drawn from 321, 85, and 68 clips, respectively, as summarized in \cref{tab:dataset_stats}. The training, validation, and test splits are subject-disjoint, so that there is no identity leakage. Following the original Gaze360 convention, we report results on three head-pose subsets defined by the absolute yaw of the target frame: \textbf{All} ($|\mathrm{yaw}|\!\le\!180^\circ$), \textbf{Semi-Front} ($|\mathrm{yaw}|\!\le\!90^\circ$), and \textbf{Front} ($|\mathrm{yaw}|\!\le\!20^\circ$); these subsets are visualized in Sec.~S1 of the supplementary material.

We also evaluate EyeTAG on the EVE dataset~\cite{park2020towards}. EVE is a large-scale video-based gaze estimation benchmark that was collected using synchronized webcam videos, screen recordings, and eye-tracking devices during natural desktop interaction scenarios. Because EVE contains temporally continuous gaze sequences, it is an appropriate benchmark for evaluating temporal autoregressive gaze estimation methods. Following the official protocol, we train on the official training split and evaluate on the official validation split, because the test annotations are not publicly available. We use only the \textit{webcam\_c} camera stream, which corresponds to the frontal webcam view. The split-level statistics of both datasets are summarized in \cref{tab:dataset_stats}.

\subsection{Experimental Setup}
\label{subsec:setup}

\noindent
\textbf{Hyperparameters.}
EyeTAG is trained and evaluated on the official training, validation, and test splits of the Gaze360~\cite{kellnhofer2019gaze360} and EVE~\cite{park2020towards} datasets. Each training sample consists of $T = 48$ consecutive frames, which provide sufficient temporal context to capture short-term head-eye dynamics. We optimize the model with AdamW using $\beta_1{=}0.9$, $\beta_2{=}0.999$, and a weight decay of $2{\times}10^{-2}$. The peak learning rate is $3{\times}10^{-4}$ with cosine annealing over 15 epochs and a 3-epoch warm-up, while pretrained backbones use a $0.1{\times}$ scaled rate. Training is performed with a batch size of 32, gradient clipping at 1.0, and a 4-frame sampling stride. Following Gaze360 preprocessing, face and eye crops are resized to $128{\times}128$.

The network is trained to minimize the angular error between the predicted gaze vector $\hat{\mathbf{g}}_t$ and the ground-truth gaze vector ${\mathbf{g}}_t$:
\begin{equation}
\mathcal{L}_{\text{ang}} = \frac{1}{P} \sum_{t=0}^{P-1}~\arccos\!\bigl(\hat{\mathbf{g}}_t \cdot \mathbf{g}_t\bigr),
\end{equation}
where $P$ is the total number of training frames. For the initial $T-1$ frames (that is, for $t=0$ to $t=T-2$), a full kinematic history is not available, so we use a padding strategy. The differential gaze history $\Delta \hat{\mathbf{G}}$ is padded with zeros, and the visual inputs (the face and eye sequences) are padded by repeating the first available frame ($\mathbf{x}_0$). In this way, the model can be trained and evaluated on the entire sequence, starting from $t=0$.

To reduce the gap between training and inference that is caused by autoregressive feedback, we use a scheduled sampling strategy when we construct the kinematic history $\Delta \hat{\mathbf{G}}$. During training, the previous gaze tokens are sampled either from the ground-truth gaze or from the cached model predictions, according to a decaying ratio $\alpha$. Specifically, $\alpha$ is linearly decreased from 1.0 at epoch 1 to 0.0 at epoch 5, so that the kinematic history gradually changes from teacher-forced supervision to fully autoregressive predictions. After epoch 5, the history is constructed entirely from the cached model predictions, which improves the robustness against accumulated prediction errors during inference. We provide a detailed ablation of this scheduled sampling strategy in Sec.~S4 of the supplementary material.

\medskip\noindent
\textbf{Model Configuration.}
The DSVE processes two visual streams: (i) a face sequence $\mathbf{X}_{t,T}^\mathrm{face}$, resized to $W_\mathrm{face} \times H_\mathrm{face} = 128 \times 128$, and (ii) bilateral eye sequences $(\mathbf{X}_{t,T}^\mathrm{L-eye},~\mathbf{X}_{t,T}^\mathrm{R-eye})$, resized to $W_\mathrm{eye} \times H_\mathrm{eye} = 128 \times 128$. The face encoder $\mathcal{F}$ uses a VGGFace2-pretrained ResNet-50~\cite{cao2018vggface2,he2016deep}, and the eye encoder $\mathcal{E}$ uses a shared ImageNet-pretrained ResNet-18~\cite{he2016deep} that is applied to each eye independently and then concatenated along the channel dimension. The ResNet-50 and ResNet-18 backbones produce 2048- and 1024-dimensional per-frame features, respectively, and the final linear layer of each visual encoder maps them to the shared dimension $D$. The TAKE $\mathcal{T}$ flattens the $T-2$ differential gaze tokens $\Delta \hat{\mathbf{G}}$ and encodes them with a two-layer MLP (hidden width $D_\mathrm{hid}=64$) whose final layer outputs the single kinematic token $\mathbf{e}^\mathrm{traj}_t$ at the shared dimension $D$. All streams use the shared latent dimension $D = 256$. The CCAF module $\mathcal{C}$ then fuses the face and eye features through cross-attention, and it models the temporal dependencies with a causal Transformer decoder ($L = 2$, $h = 4$). Finally, the 3D gaze vector $\hat{\mathbf{g}}_t$ is regressed from the last output token $\mathbf{h}_{t}$.

We measure inference latency on a single NVIDIA RTX A6000 GPU in FP32 with a batch size of 1. Under this desktop-GPU setting, EyeTAG requires approximately 20 ms per $T=48$ input window, which is below the 33.3 ms frame interval of 30 Hz video. This result should be interpreted as feasibility on the evaluated hardware rather than as a claim of embedded real-time deployment.

\medskip\noindent
\textbf{Evaluation metric.}
We measure the model performance using the Mean Angular Error (MAE), which is computed in degrees as:
\begin{equation}
\mathrm{MAE} = \frac{1}{P_\mathrm{test}}\sum_{t=0}^{P_\mathrm{test}-1}
\arccos\!\left(
\frac{\hat{\mathbf{g}}_t \cdot \mathbf{g}_t}{
\|\hat{\mathbf{g}}_t\|_2 \|\mathbf{g}_t\|_2}
\right) \!\times \!\frac{180}{\pi},
\label{eq:angular_error}
\end{equation}
where $P_\mathrm{test}$ is the total number of frames in the test set. As in training, we evaluate all frames and use the same padding strategy for the initial $T-1$ frames. A smaller MAE means higher accuracy. This metric measures the angular difference between the predicted and the ground-truth gaze directions, and it is a standard metric for 3D gaze estimation.

\begin{table}[t!]
\centering
\footnotesize
\setlength{\tabcolsep}{2pt}
\resizebox{\textwidth}{!}{
\begin{tabular}{llcccc@{\hspace{1.5em}}c@{\hspace{1.5em}}cc}
\toprule
\multirow{2}{*}{\textbf{Type}} & \multirow{2}{*}{\textbf{Method}} & \multicolumn{4}{c}{\textbf{(a) Gaze360 Dataset~\cite{kellnhofer2019gaze360}}} & \multicolumn{1}{c}{\textbf{(b) EVE Dataset~\cite{park2020towards}}} & \multicolumn{2}{c}{\textbf{(c) Complexity}} \\
\cmidrule(lr){3-6}\cmidrule(lr){7-7}\cmidrule(lr){8-9}
 & & \textbf{All} & \textbf{Semi-Front} & \textbf{Front} & \textbf{Mean} & \textbf{All} & \textbf{Time (ms)} & \textbf{GFLOPs} \\
\midrule
\multirow{3}{*}{SF} & L2CS-Net~\cite{abdelrahman2023l2cs} & 11.10 $\pm$ 9.82 & 10.77 $\pm$ 8.30 & 9.92 $\pm$ 7.97 & 10.60 $\pm$ 8.70 & 5.06 $\pm$ 6.12 & 5.42 & 16.57 \\
 & GazeTR-Hybrid~\cite{cheng2022gaze} & 10.56 $\pm$ 9.24 & 10.30 $\pm$ 8.05 & 9.63 $\pm$ 7.90 & 10.16 $\pm$ 8.40 & 3.63 $\pm$ 3.14 & \underline{5.35} & \underline{1.84} \\
 & CrossGaze~\cite{catruna2024crossgaze} & 10.47 $\pm$ 8.45 & 10.29 $\pm$ 7.79 & \underline{9.23 $\pm$ 7.31} & 10.00 $\pm$ 7.85 & 3.30 $\pm$ 2.68 & 14.31 & \textbf{1.43} \\
\midrule
\multirow{3}{*}{MF} & Gaze360~\cite{kellnhofer2019gaze360} & \underline{10.46 $\pm$ 8.92} & \underline{10.18 $\pm$ 7.69} & 9.32 $\pm$ 7.58 & \underline{9.99 $\pm$ 8.06} & 2.73 $\pm$ 1.77 & \textbf{2.83} & 12.79 \\
 & STAGE~\cite{jindal2024spatio} & 12.08 $\pm$ 10.82 & 11.63 $\pm$ 8.53 & 9.43 $\pm$ 6.72 & 11.05 $\pm$ 8.69 & \underline{2.58 $\pm$ 2.11} & 10.26 & 39.55 \\
 & \textbf{EyeTAG (Ours)} & \textbf{9.29 $\pm$ 7.78} & \textbf{9.12 $\pm$ 7.23} & \textbf{8.08 $\pm$ 7.23} & \textbf{8.83 $\pm$ 7.41} & \textbf{2.56 $\pm$ 1.89} & 19.97 & 121.57 \\
\bottomrule
\end{tabular}
}
\caption{Quantitative comparison of 3D gaze estimation on the (a) Gaze360 and (b) EVE datasets, together with (c) the model complexity. We report the Mean Angular Error (MAE, \textdegree; lower is better), where the \textbf{Mean} column is the average over the All, Semi-Front, and Front subsets. The (c) Complexity columns report the inference time (in milliseconds) and the computational cost (GFLOPs). SF and MF denote single-frame and multi-frame methods. The \textbf{best} and \underline{second-best} results in each column are marked in bold and underline.}
\label{tab:main_results}
\end{table}

\subsection{Experimental Results}
\label{subsec:results}

\Cref{tab:main_results} summarizes the quantitative performance on the Gaze360~\cite{kellnhofer2019gaze360} and EVE~\cite{park2020towards} benchmarks. We compare EyeTAG with two groups of baselines: single-frame methods (L2CS-Net~\cite{abdelrahman2023l2cs}, GazeTR-Hybrid~\cite{cheng2022gaze}, and CrossGaze~\cite{catruna2024crossgaze}) and multi-frame methods (Gaze360~\cite{kellnhofer2019gaze360} and STAGE~\cite{jindal2024spatio}). All baseline numbers are obtained from our re-training under the same evaluation protocol. STAGE is evaluated in the same calibration-free setting as EyeTAG, although it was originally designed for personalized gaze estimation; this difference likely contributes to its lower Gaze360 score. We also report the (c) model complexity, namely the average inference time and the computational cost (GFLOPs), measured on a single NVIDIA RTX A6000 (FP32, batch size~1).

On Gaze360, EyeTAG consistently achieves the lowest MAE on all head-pose subsets, which shows that it is effective for both frontal and large-pose variations. Specifically, EyeTAG achieves 9.29\textdegree{} on the full-range subset, 9.12\textdegree{} on the semi-frontal subset, and 8.08\textdegree{} on the frontal subset, and it improves the MAE by about 1.0\textdegree{} over the strongest baseline.

To contextualize the source of this gain, we also performed a backbone-controlled check using the same VGGFace2 ResNet-50 face encoder as EyeTAG. This improves the strongest multi-frame baseline from 9.99\textdegree{} to 9.76\textdegree{}, whereas EyeTAG remains at 8.83\textdegree{}; together with the visual-only EyeTAG result in \cref{tab:gaze_effect} (9.19\textdegree{}), this indicates that the improvement comes from the overall causal fusion design and the differential prior, not from the face encoder alone.

We observe a similar trend on EVE, where EyeTAG achieves a mean MAE of 2.56\textdegree{}, remaining on par with the strongest multi-frame baselines (STAGE 2.58\textdegree{} and Gaze360 2.73\textdegree{}) while yielding the lowest mean error overall. The gap among these top methods is small on this near-frontal benchmark, whereas the single-frame methods show noticeably higher errors (CrossGaze 3.30\textdegree{} and GazeTR-Hybrid 3.63\textdegree{}). EyeTAG also shows a lower error standard deviation than STAGE ($1.89$ vs. $2.11$). Although Gaze360 reports a slightly smaller deviation ($1.77$), it relies on future frames when estimating the current gaze and is therefore not causal. Among methods restricted to past and current observations, EyeTAG provides a favorable balance of accuracy and prediction consistency. Since the mean MAE is nearly saturated on EVE and does not fully reflect frame-to-frame behavior, we further analyze temporal stability in the following section.

\begin{table}[t!] \centering \footnotesize \setlength{\tabcolsep}{3pt} \begin{tabular}{llccc} \toprule \textbf{Type} & \textbf{Method} & \textbf{(a) Fixation Jitter} & \textbf{(b) Saccade Bias} & \textbf{(c) Absolute Mean} \\ \midrule \multirow{3}{*}{SF} & L2CS-Net~\cite{abdelrahman2023l2cs} & 5.60 $\pm$ 0.16 & \textbf{0.02 $\pm$ 0.17} & 2.81 $\pm$ 0.17 \\ & GazeTR-Hybrid~\cite{cheng2022gaze} & 5.48 $\pm$ 0.16 & 0.84 $\pm$ 0.18 & 3.16 $\pm$ 0.17 \\ & CrossGaze~\cite{catruna2024crossgaze} & 5.06 $\pm$ 0.14 & 0.23 $\pm$ 0.15 & 2.65 $\pm$ 0.15 \\ \midrule \multirow{5}{*}{MF} & Gaze360~\cite{kellnhofer2019gaze360} & \textbf{1.71 $\pm$ 0.10} & -4.24 $\pm$ 0.17 & 2.98 $\pm$ 0.14 \\ & STAGE~\cite{jindal2024spatio} & 3.81 $\pm$ 0.11 & -0.89 $\pm$ 0.26 & 2.35 $\pm$ 0.19 \\ \cmidrule(lr){2-5} & EyeTAG (Visual-only) & 3.51 $\pm$ 0.12 & -0.46 $\pm$ 0.16 & 1.99 $\pm$ 0.14 \\ & EyeTAG (Normal prior) & \underline{3.43 $\pm$ 0.09} & -0.53 $\pm$ 0.14 & \underline{1.98 $\pm$ 0.12} \\ & \textbf{EyeTAG (Ours)} & 3.68 $\pm$ 0.12 & \underline{0.05 $\pm$ 0.15} & \textbf{1.87 $\pm$ 0.14} \\ \bottomrule \end{tabular} \caption{Temporal behavior analysis (all values in \textdegree/frame). (a) \textbf{Fixation Jitter} $J$: mean predicted velocity under near-static gaze ($v^{\mathrm{GT}}<1^\circ$/frame, $N{=}1{,}003$ frame pairs); lower is better. (b) \textbf{Saccade Bias} $B$: mean signed velocity difference $v^{\mathrm{pred}}-v^{\mathrm{GT}}$ under rapid gaze transitions ($v^{\mathrm{GT}}>5^\circ$/frame, $N{=}3{,}306$ frame pairs); closer to zero is better. (c) \textbf{Absolute Mean} $\tfrac{1}{2}(J{+}|B|)$; lower is better. SF and MF denote single-frame and multi-frame methods, respectively. The EyeTAG variants keep the encoder and remaining architecture fixed and differ only in the gaze-history component. The \textbf{best} and \underline{second-best} results in each column are marked in bold and underline, respectively.} \label{tab:temporal_behavior} \end{table}

\Cref{tab:temporal_behavior} reports our temporal stability analysis, including two controlled EyeTAG variants that keep the encoder and architecture fixed while changing only the gaze-history component. We define the frame-to-frame angular velocities of the ground-truth and predicted gaze as \begin{equation} v^{\mathrm{GT}}_t = \arccos\!\left(\mathbf{g}_{t-1}\!\cdot\!\mathbf{g}_t\right) \!\cdot\!\tfrac{180}{\pi}, \qquad v^{\mathrm{pred}}_t = \arccos\!\left(\hat{\mathbf{g}}_{t-1}\!\cdot\!\hat{\mathbf{g}}_t\right) \!\cdot\!\tfrac{180}{\pi}, \label{eq:gaze_velocity} \end{equation} which measure the angular displacement between consecutive gaze directions. Using these velocities, we evaluate prediction stability during near-static gaze and responsiveness during rapid gaze transitions.

We first consider \emph{fixation jitter} under near-static gaze ($v^{\mathrm{GT}}_t < 1^\circ$/frame). For these frame pairs, we report $J = \mathbb{E}[v^{\mathrm{pred}}_t \mid v^{\mathrm{GT}}_t < 1^\circ]$, which measures prediction fluctuation when the true gaze is nearly stationary. Gaze360 shows the lowest jitter, likely aided by its use of future frames around the target. Among the EyeTAG variants, the kinematic prior keeps jitter in a similar range to the visual-only and normal-prior variants ($3.68$ vs.\ $3.51$ and $3.43$), indicating that its benefit does not arise from additional temporal smoothing.

We next consider \emph{saccade responsiveness} under rapid gaze motion ($v^{\mathrm{GT}}_t > 5^\circ$/frame). We report the signed velocity difference $B = \mathbb{E}[v^{\mathrm{pred}}_t - v^{\mathrm{GT}}_t \mid v^{\mathrm{GT}}_t > 5^\circ]$, where values closer to zero are better. Negative values indicate systematic under-shooting, or over-smoothing. Without the kinematic prior, both the visual-only and normal-prior variants under-shoot saccades ($B=-0.46$ and $-0.53$), whereas the full EyeTAG reduces this bias to $0.05$. Since the encoder and remaining architecture are fixed across these variants, this comparison directly attributes the improved saccade responsiveness to the explicit kinematic prior rather than general smoothing or the encoder.

Finally, we summarize both behaviors using $M=\tfrac{1}{2}(J+|B|)$. EyeTAG achieves the lowest $M$ of $1.87$, compared with $1.99$ and $1.98$ for the visual-only and normal-prior variants, respectively, demonstrating the best overall balance between fixation stability and responsiveness.

\begin{figure*}[t!]
  \centering
  \includegraphics[width=\linewidth]{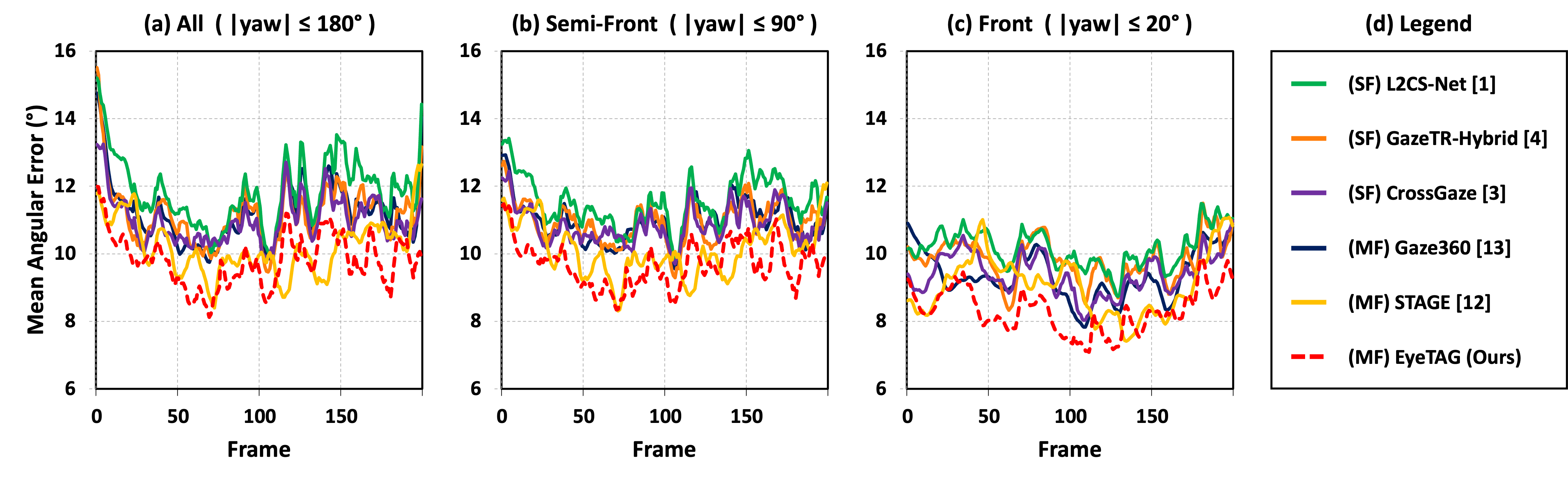}
  \vspace{-25pt}
  \caption{Per-frame Mean Angular Error (MAE, \textdegree) on Gaze360 for the (a)~All, (b)~Semi-Front, and (c)~Front head-pose subsets (lower is better). }
  \label{fig:perframe}
\end{figure*}

Beyond these aggregate metrics, \cref{fig:perframe} traces the mean angular error frame by frame along the test sequence. EyeTAG (red, dashed) generally remains lower than the baselines across the sequence on every head-pose subset, which shows that its accuracy advantage is consistent over time rather than an artifact of averaging.

\section{Ablation Studies}

\begin{table}[t]
\centering
\footnotesize
\setlength{\tabcolsep}{3pt}
\begin{tabular}{c c c c c c}
\toprule
& \textbf{Fusion Strategy} & \textbf{All} & \textbf{Semi-Front} & \textbf{Front} & \textbf{Mean} \\
\midrule
(a) & Self-Attn + Concat (face $\parallel$ eye) & 9.80 $\pm$ 8.14 & 9.60 $\pm$ 7.45 & 8.71 $\pm$ 7.43 & 9.37 $\pm$ 7.67 \\
(b) & Bi. Cross-Attn (face $\leftrightarrow$ eye) & \underline{9.66 $\pm$ 8.16} & \underline{9.49 $\pm$ 7.55} & \underline{8.62 $\pm$ 7.49} & \underline{9.26 $\pm$ 7.73} \\
\midrule
\textbf{(Ours)} & Uni. Cross-Attn (face $\rightarrow$ eye)
& \textbf{9.29 $\pm$ 7.78}
& \textbf{9.12 $\pm$ 7.23}
& \textbf{8.08 $\pm$ 7.23}
& \textbf{8.83 $\pm$ 7.41} \\
\bottomrule
\end{tabular}
\caption{Effect of fusion strategy. MAE (\textdegree); lower is better. The \textbf{best} and \underline{second-best} results in each column are marked in bold and underline, respectively.}
\label{tab:ablation_fusion}
\end{table}

\subsection{Effect of Cross-Modal Fusion Strategy}
\label{subsec:ablation_fusion}

We study how to effectively fuse the face and eye embeddings from the visual encoders. We compare three fusion strategies in \cref{tab:ablation_fusion}. \textbf{Self-Attn + Concat} processes the face and eye embeddings separately, concatenates them, and projects the result back to the original dimension. \textbf{Bidirectional Cross-Attn} applies cross-attention in both the face-to-eye and the eye-to-face directions before projection. Our proposed \textbf{Unidirectional Cross-Attn} uses the face embeddings as queries and the eye embeddings as keys and values.

As shown in \cref{tab:ablation_fusion}, the proposed unidirectional cross-attention achieves the lowest error on all subsets. Compared with self-attention and bidirectional fusion, it injects the fine-grained eye cues into the face representation more effectively, without redundant fusion paths. This shows that the face embedding should be the main contextual representation, while the eye features provide complementary gaze-specific cues.

\subsection{Influence of Gaze History Embedding}
\label{subsec:ablation_prevgaze}

\begin{table}[t]
\centering
\footnotesize
\setlength{\tabcolsep}{3pt}
\begin{tabular}{c c c c c c}
\toprule
& \textbf{Gaze History} & \textbf{All} & \textbf{Semi-Front} & \textbf{Front} & \textbf{Mean} \\
\midrule
(a) & No Gaze History (visual-only) & 9.72 $\pm$ 8.15 & 9.55 $\pm$ 7.76 & 8.30 $\pm$ 7.53
& 9.19 $\pm$ 7.81 \\
(b) & Normal Gaze Prior ($w/~\hat{G}$) & \underline{9.63 $\pm$ 7.94} & \underline{9.46 $\pm$ 7.38}
& \underline{8.22 $\pm$ 7.07} & \underline{9.10 $\pm$ 7.46} \\
\midrule
\textbf{(Ours)} & Kinematic Gaze Prior ($w/~\Delta\hat{G}$)
& \textbf{9.29 $\pm$ 7.78}
& \textbf{9.12 $\pm$ 7.23}
& \textbf{8.08 $\pm$ 7.23}
& \textbf{8.83 $\pm$ 7.41} \\
\bottomrule
\end{tabular}
\caption{Ablation study on the gaze history representation. MAE (\textdegree); lower is better. The proposed kinematic prior $\Delta\hat{\mathbf{G}}$ achieves the lowest error across all subsets. The \textbf{best} and \underline{second-best} results in each column are marked in bold and underline, respectively.}
\label{tab:gaze_effect}
\end{table}

We further study how the representation of the gaze history affects temporal reasoning. We evaluate three variants in \cref{tab:gaze_effect}: (a) \textbf{No Gaze History} (\textbf{visual-only}), (b) \textbf{Normal Gaze Prior} ($\textbf{w/}~\hat{\mathbf{G}}$), and (c) our proposed \textbf{Kinematic Gaze Prior} ($\textbf{w/}~\Delta\hat{\mathbf{G}}$).

As summarized in \cref{tab:gaze_effect}, removing the prior entirely (visual-only) gives the highest error on all subsets, the Normal Gaze Prior ($\textbf{w/}~\hat{\mathbf{G}}$) consistently improves over it, and our Kinematic Gaze Prior ($\textbf{w/}~\Delta\hat{\mathbf{G}}$) performs best across all subsets, supporting the benefit of explicitly modeling gaze dynamics.

\begin{wrapfigure}[12]{r}{0.55\linewidth}
    \centering
    \vspace{-15pt}
    \includegraphics[width=\linewidth]{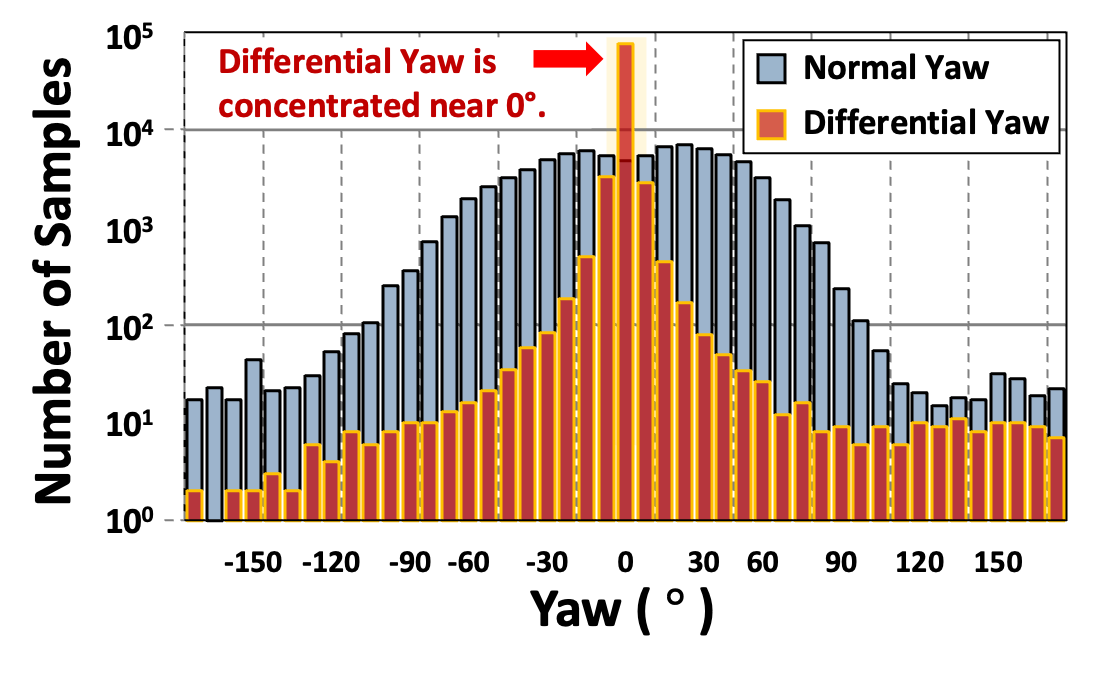}
    \vspace{-30pt}
    \caption{Distribution of normal versus differential gaze yaw on Gaze360. The $y$-axis is on a log scale.}
    \label{fig:graph}
    \vspace{-8pt}
\end{wrapfigure}
The Normal Gaze Prior $\hat{\mathbf{G}}$ is sensitive to subject-specific static biases, such as a person's default resting gaze. In contrast, the Kinematic Gaze Prior $\Delta \hat{\mathbf{G}}$ encodes relative motion, such as velocity and smoothness, so it encourages the network to learn a more general representation of gaze dynamics.
As shown in \cref{fig:graph}, the normal yaw distribution covers a broad $\pm180^{\circ}$ range with large variance ($\sigma\!\approx\!39^\circ$), while the differential yaw forms a compact, zero-centered distribution with much lower variance ($\sigma\!\approx\!12^\circ$). This compact representation preserves the shared motion dynamics and reduces subject-dependent gaze offsets, which helps the network distinguish genuine gaze transitions from noise. We further show in Sec.~S6 of the supplementary material that the kinematic prior remains effective even in the early frames of a clip, where the gaze history is still incomplete ($t < T$).

Finally, to bound the cost of autoregressive feedback, we re-evaluated the model with the gaze history replaced by the ground truth. This oracle improves the mean MAE only from 8.83\textdegree{} to 8.65\textdegree{}, and the gap stays flat across clip positions (0.16--0.24\textdegree{} for frames 0--29, 30--99, 100--199 and 200+), so scheduled sampling keeps the autoregressive error from accumulating over time.

\subsection{Effect of Temporal Window Length}
\label{subsec:frame-length}

To study the influence of temporal context on gaze estimation, we change the window length $T\in\{4, ~8, ~16, ~24, ~32, ~48, ~64\}$ under the same training and testing conditions. As reported in \cref{tab:temporal_inference_time}, the MAE decreases steadily as $T$ increases from 4 to 48 frames, which shows that a longer temporal context improves stability and accuracy. However, the performance saturates after $T=48$ and changes only slightly at $T=64$, while the inference time and the computational cost keep increasing.

We also report the measured latency in \cref{tab:temporal_inference_time}. On the RTX A6000, processing one input window takes 19.97 ms at $T=48$ and 25.23 ms at $T=64$, which is below the 33.3 ms frame interval of 30 Hz video on this hardware. This supports desktop-GPU feasibility, but resource-constrained deployment would require lighter encoders, feature caching, or other optimization. The inference time is also not proportional to the GFLOPs: for $T\!\le\!32$ it stays around 11--14 ms, where fixed overhead dominates the latency. We therefore adopt $T=48$ as the default window for its accuracy--latency balance. In the Normal Gaze Prior variant ($\textbf{w/}~\hat{\mathbf{G}}$), the best MAE is achieved at $T=32$, the detailed results of which are reported in Sec.~S5 of the supplementary material.

\begin{table}[t]
\centering
\footnotesize
\setlength{\tabcolsep}{3pt}
\begin{tabular}{lccccccc}
\toprule
$T$ & 4 & 8 & 16 & 24 & 32 & \textbf{48}$^{\ast}$ & 64 \\
\midrule
\textbf{Mean} & 9.87 $\pm$ 7.57 & 9.49 $\pm$ 7.36 & 9.47 $\pm$ 7.41 & 9.35 $\pm$ 7.56 & 8.91 $\pm$ 7.39 & \underline{8.83 $\pm$ 7.41} & \textbf{8.82 $\pm$ 7.40} \\
\textbf{Time (ms)} & 10.78 & 11.12 & 11.38 & 11.41 & 13.91 & 19.97 & 25.23 \\
\textbf{GFLOPs} & 10.13 & 20.26 & 40.53 & 60.79 & 81.05 & 121.57 & 162.09 \\
\bottomrule
\end{tabular}

\caption{Effect of temporal window size $T$. For each $T$ we report the Mean Angular Error (MAE, \textdegree), the average inference time (Time, in milliseconds) per input window on a single NVIDIA RTX A6000 (FP32, batch size~1), and the computational cost (GFLOPs); $^{\ast}$ denotes the adopted setting. Accuracy improves up to $T=48$ and then saturates, while the inference time and computational cost keep increasing. The \textbf{best} and \underline{second-best} Mean values are marked in bold and underline, respectively.}
\label{tab:temporal_inference_time}
\end{table}

\section{Conclusion}
\label{sec:conclusion}
In this work, we present EyeTAG, a trajectory-aware multi-frame gaze estimation framework that addresses the temporal instability of frame-based approaches. Unlike appearance-only methods that predict gaze independently for each frame, EyeTAG explicitly models gaze kinematics through differential gaze trajectories, so that the prediction becomes temporally coherent and stable under unconstrained head-eye motion. It integrates multi-scale appearance cues from the face and eye regions with a trajectory-aware kinematic prior learned from the differential gaze dynamics $\Delta \hat{\mathbf{G}}$, using cross-modal attention fusion and causal temporal aggregation. On Gaze360, EyeTAG clearly reduces angular error, and on EVE it remains on par with the strongest multi-frame baseline. It also keeps fixation jitter low and reduces velocity bias during rapid transitions, and within-model ablations attribute this near-zero saccade bias to the differential prior rather than to temporal context alone.

\medskip\noindent
\textbf{Limitations.} The current framework only estimates the individual gaze direction, and it does not use scene context. EyeTAG models the temporal dynamics of \textit{where} a person is looking, but it does not reason about \textit{what} the person is looking at, or about how multiple people's gaze patterns interact in a shared environment. Real-time performance is verified only on a desktop GPU, so embedded deployment remains future work. Bridging this gap between low-level gaze estimation and higher-level attention understanding is an important direction for future research.

\medskip\noindent
\textbf{Future Work.} We plan to extend this framework to multi-person and scene-aware gaze understanding, such as social gaze interactions and contextual intent, moving toward next-generation systems for complex human attention.

\medskip\noindent
\textbf{Acknowledgements.}
This research was supported by G-LAMP Program of the National Research Foundation of Korea (NRF) grant funded by the Ministry of Education (No. RS-2025-25441317), by the IITP (Institute of Information \& Communications Technology Planning \& Evaluation)-ITRC (Information Technology Research Center) grant funded by the Korean government (Ministry of Science and ICT) (IITP-2026-RS-2020-II201602), by the National Research Foundation of Korea (NRF) grant funded by the Korean government (MSIT) (RS-2025-24533373), and by the Cyber Investigation Support Technology Development Program (No. RS-2025-02304983) of the Korea Institute of Police Technology (KIPoT), funded by the Korean National Police Agency.

\bibliography{egbib}

\clearpage
\setcounter{section}{0}
\setcounter{figure}{0}
\setcounter{table}{0}
\setcounter{equation}{0}
\renewcommand{\thesection}{S\arabic{section}}
\renewcommand{\thefigure}{S\arabic{figure}}
\renewcommand{\thetable}{S\arabic{table}}
\renewcommand{\thealgorithm}{S\arabic{algorithm}}
\renewcommand{\theHsection}{S\arabic{section}}
\renewcommand{\theHfigure}{S\arabic{figure}}
\renewcommand{\theHtable}{S\arabic{table}}
\renewcommand{\theHalgorithm}{S\arabic{algorithm}}
\begin{center}
    {\LARGE\bfseries Supplementary Material\par}
    \vspace{0.2cm}
    {\Large\bfseries EyeTAG: Eye Trajectory-Aware Gaze Estimation\par}
\end{center}

\vspace{0.5cm}
\section{Dataset Details}
\label{sec:supp-dataset}

\begin{figure}[H]
    \centering
    \includegraphics[width=0.4\linewidth]{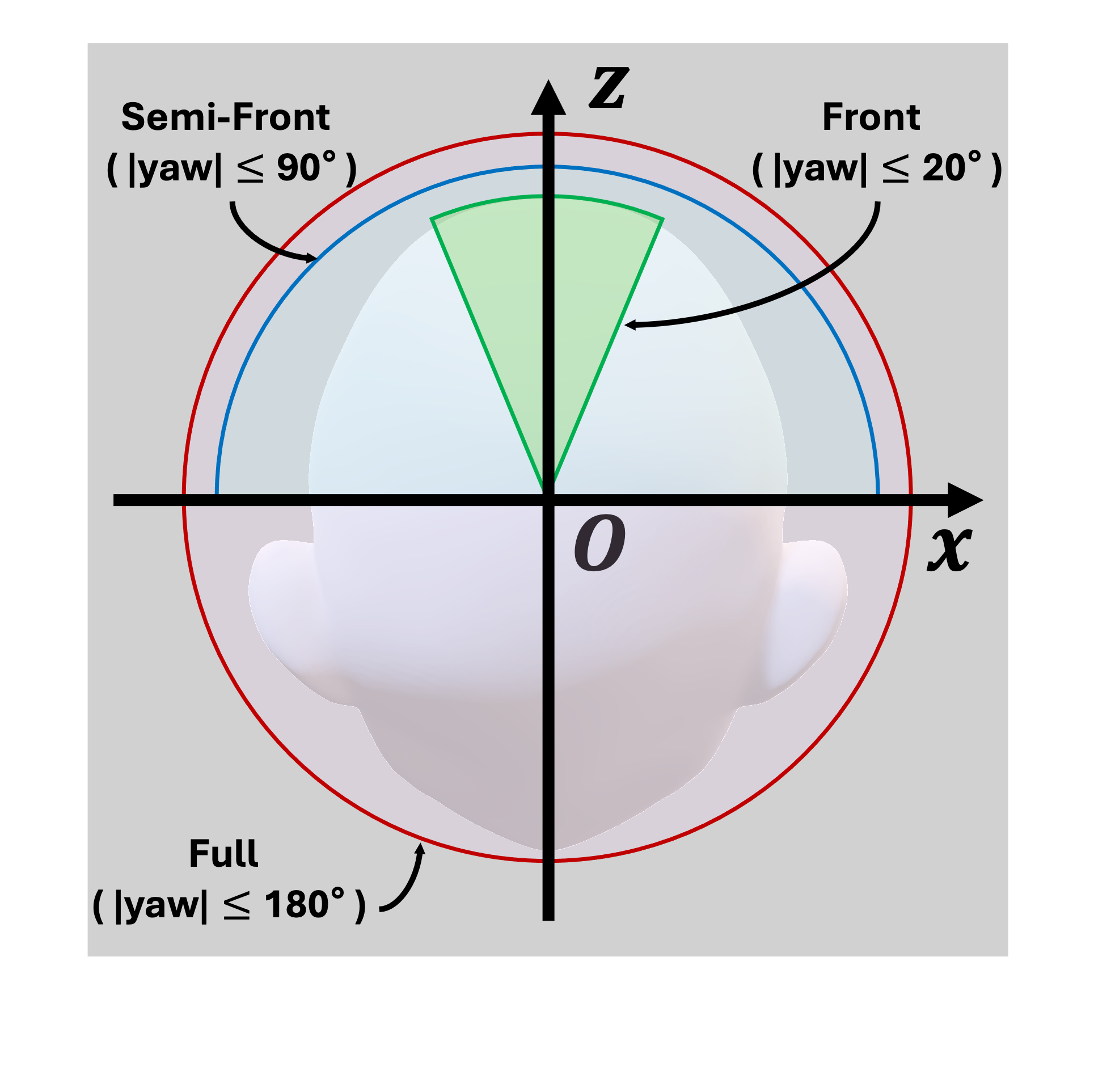}
    \vspace{-20pt}
    \caption{
        Gaze360 evaluation subsets on the unit sphere:
        \textbf{Front} (green, $|\mathrm{yaw}| \leq 20^\circ$),
        \textbf{Semi-Front} (blue, $|\mathrm{yaw}| \leq 90^\circ$), and
        \textbf{All} (red, $|\mathrm{yaw}| \leq 180^\circ$).
    }
    \label{fig:supp_subset}
\end{figure}

\noindent\textbf{Gaze360 Subset Visualization.} The main paper defines the three evaluation subsets in the text. Here, \cref{fig:supp_subset} shows their spatial coverage on the unit sphere. The \textbf{Front} subset ($|\mathrm{yaw}| \leq 20^\circ$) is a narrow frontal cone, where the facial and ocular appearance is most informative. The \textbf{Semi-Front} subset ($|\mathrm{yaw}| \leq 90^\circ$) covers the full forward-facing hemisphere. The \textbf{All} subset ($|\mathrm{yaw}| \leq 180^\circ$) covers the entire sphere, including backward-facing directions.

\section{Effect of Face-Encoder Pretraining}
\label{sec:supp-pretraining}

A key design choice in our Dual-Stream Visual Encoder (DSVE) is how the face encoder is pretrained. To study this effect alone, we fix the face-encoder architecture to ResNet-50 and the eye encoder to ResNet-18, and we change only the pretraining of the face encoder across three settings: training \emph{from scratch} (no pretraining), generic \emph{ImageNet} pretraining~\cite{deng2009imagenet}, and face-domain \emph{VGGFace2} pretraining~\cite{cao2018vggface2}. Because all the other components are fixed, any difference in performance comes from the pretraining strategy, not from the model capacity.

As shown in \cref{tab:backbone_selection}, the accuracy improves consistently as the pretraining becomes more domain-relevant.

Training from scratch yields the largest error, likely because the gaze dataset alone is insufficient for learning robust facial representations. Generic ImageNet pretraining gives a clear improvement, because it provides transferable low- and mid-level visual features. Face-domain VGGFace2 pretraining performs best, and it reduces the mean angular error to 8.83\textdegree{}, with consistent gains across the All, Semi-Front, and Front subsets. This advantage comes from the identity- and pose-aware facial features learned from large-scale face data. These features transfer better than generic ImageNet features to the extreme head poses and diverse identities in Gaze360~\cite{kellnhofer2019gaze360}.

These results show that face-domain pretraining is important for in-the-wild gaze estimation, because the head pose and the subjects vary widely. At the same time, the gains from VGGFace2 pretraining are separate from the architectural contributions of EyeTAG. The cross-modal fusion and gaze-history ablations in the main paper keep the VGGFace2 backbone fixed, so they isolate the contribution of the proposed architecture. In the same way, the temporal-stability analysis mainly reflects the kinematics-aware design, not the choice of backbone initialization. We therefore include this experiment as supplementary evidence to clarify the respective contributions of backbone pretraining and the proposed EyeTAG framework.

\begin{table}[t]
\centering
\footnotesize
\setlength{\tabcolsep}{4pt}
\begin{tabular}{clcccc}
\toprule
& \textbf{Pretraining} & \textbf{All} & \textbf{Semi-Front} & \textbf{Front} & \textbf{Mean} \\
\midrule
(a) & From scratch & 20.41 $\pm$ 16.39 & 18.03 $\pm$ 15.54 & 16.05 $\pm$ 14.50 & 18.16 $\pm$ 15.47 \\
(b) & ImageNet~\cite{deng2009imagenet} & \underline{11.69 $\pm$ 10.22} & \underline{11.34 $\pm$ 8.64} & \underline{10.15 $\pm$ 8.50} & \underline{11.06 $\pm$ 9.12} \\
\midrule
(\textbf{Ours}) & VGGFace2~\cite{cao2018vggface2} & \textbf{9.29 $\pm$ 7.78} & \textbf{9.12 $\pm$ 7.23} & \textbf{8.08 $\pm$ 7.23} & \textbf{8.83 $\pm$ 7.41} \\
\bottomrule
\end{tabular}
\caption{Effect of the face encoder $\mathcal{F}$ pretraining strategy on Gaze360~\cite{kellnhofer2019gaze360}. The face encoder architecture (ResNet-50) and the eye encoder (ResNet-18) are held fixed. Only the pretraining of the face encoder is varied across three settings: from scratch, generic ImageNet pretraining, and face-domain VGGFace2 pretraining. MAE (\textdegree); lower is better. The \textbf{best} and \underline{second-best} results in each column are marked in bold and underline, respectively.}
\label{tab:backbone_selection}
\end{table}

\section{Effect of Eye-Encoder Backbone}
\label{sec:supp-eye-backbone}

The main paper uses a shared ImageNet-pretrained ResNet-18 as the eye encoder $\mathcal{E}$. It processes each frame independently, without any temporal modeling inside the eye branch. Here we test whether a backbone with stronger spatial capacity or with explicit temporal modeling gives a further improvement. All experiments use the same configuration as our model ($T{=}48$, kinematic gaze prior, VGGFace2-pretrained face encoder), and we change only the eye backbone.

\begin{table}[t]
\centering
\footnotesize
\setlength{\tabcolsep}{4pt}
\begin{tabular}{llcccc}
\toprule
\textbf{Eye Backbone} & \textbf{Type} & \textbf{All} & \textbf{Semi-Front} & \textbf{Front} & \textbf{Mean} \\
\midrule
EfficientNet-B0 \cite{tan2019efficientnet} & 2D, no temporal & 10.77 $\pm$ 10.80 & 10.37 $\pm$ 8.83 & 8.26 $\pm$ 6.56 & 9.80 $\pm$ 8.73 \\
TSM \cite{lin2019tsm} & 2D + temporal & 9.47 $\pm$ 7.89 & 9.30 $\pm$ 7.30 & 8.31 $\pm$ 7.35 & 9.03 $\pm$ 7.51 \\
3D CNN \cite{tran2015learning}(scratch) & 3D, temporal  & \underline{9.41 $\pm$ 7.90} & \underline{9.22 $\pm$ 7.27} & \underline{8.17 $\pm$ 7.06} & \underline{8.93 $\pm$ 7.41} \\
\midrule
(\textbf{Ours}) \textbf{ResNet-18 (shared)} & 2D, no temporal & \textbf{9.29 $\pm$ 7.78} & \textbf{9.12 $\pm$ 7.23} & \textbf{8.08 $\pm$ 7.23} & \textbf{8.83 $\pm$ 7.41} \\
\bottomrule
\end{tabular}
\caption{
    Effect of the eye-encoder backbone on Gaze360~\cite{kellnhofer2019gaze360}. MAE (\textdegree); lower is better. The \textbf{best} and \underline{second-best} results in each column are marked in bold and underline, respectively.
}
\label{tab:supp_eye_backbone}
\end{table}

As shown in \cref{tab:supp_eye_backbone}, adding explicit temporal modeling to the eye branch does not improve the performance. Both TSM and the 3D CNN are worse than the simple 2D ResNet-18 baseline, and replacing ResNet-18 with the stronger EfficientNet-B0 also gives no improvement. This suggests that at $T{=}48$ the downstream CCAF module already captures enough ocular dynamics from the long temporal context, so extra temporal modeling or stronger spatial capacity inside the eye encoder is redundant.

\section{Effect of Scheduled Sampling}
\label{sec:supp-scheduled-sampling}

During training, the gaze history $\hat{\mathbf{G}}_{t-1,T-1}$ fed to the TAKE module can be constructed from the ground-truth gaze (teacher forcing) or from the model's own past predictions (autoregressive). To reduce the train-inference mismatch between these two strategies, we use a scheduled sampling strategy~\cite{bengio2015scheduled} that linearly anneals the teacher-forcing ratio $\alpha$ from 1.0 to 0.0 over epochs 1-5. As summarized in \Cref{alg:eyetag}, after each epoch the current model runs inference on the entire training set, and the resulting predictions are stored in the cache $\mathcal{C}$ for use in the next epoch. At inference, the model processes frames sequentially, appending each prediction to $\mathcal{C}$ online. When fewer than $T{-}1$ past predictions are available, the cache is padded by repeating the first prediction $\hat{\mathbf{g}}_1$.

\begin{algorithm}[t]
\centering
\footnotesize
\begin{tabular}{p{0.92\linewidth}}
\hline
\textbf{Input:} $\mathbf{X}_{t,T} = \{\mathbf{x}_{t-T+1},~\cdots,~\mathbf{x}_{t}\}$, window size $T$ \\
\textbf{Output:} $\hat{\mathbf{g}}_t$ \\
\hline
\textit{--- Training (per epoch) ---} \\
\quad 1.\ \textbf{for each} frame $t$ \textbf{do} \\
\quad\quad a.\ $\mathbf{e}_t^{\mathrm{face}},~\mathbf{e}_t^{\mathrm{eye}} \leftarrow \mathrm{DSVE}(\mathbf{X}_{t,T})$ \\
\quad\quad b.\ Construct gaze history: \\
\quad\quad\quad\quad sample $\hat{\mathbf{G}}_{t-1,T-1}$ from $\mathbf{G}^{\mathrm{GT}}_{t-1,T-1}$ or $\mathcal{C}$ with ratio $\alpha$ \hfill $\triangleright$ $\alpha: 1.0 \rightarrow 0.0$ over ep.\ 1--5 \\
\quad\quad c.\ $\Delta\hat{\mathbf{G}}_{t-1,T-1} \leftarrow [\hat{\mathbf{g}}_{t-k} - \hat{\mathbf{g}}_{t-k-1}]_{k=1}^{T-2}$ \\
\quad\quad d.\ $\mathbf{e}_t^{\mathrm{traj}} \leftarrow \mathrm{TAKE}(\Delta\hat{\mathbf{G}}_{t-1,T-1})$ \\
\quad\quad e.\ $\hat{\mathbf{g}}_t \leftarrow \mathrm{GVP}\!\left(\mathrm{CCAF}(\mathbf{e}_t^{\mathrm{face}},~\mathbf{e}_t^{\mathrm{eye}},~\mathbf{e}_t^{\mathrm{traj}})\right)$ \\
\quad 2.\ After epoch ends: run inference on entire training set $\rightarrow$ store in $\mathcal{C}$ \\
\hline
\textit{--- Inference (per frame) ---} \\
\quad 1.\ Retrieve gaze history from $\mathcal{C}$ \\
\quad\quad \textbf{if} $|\mathcal{C}| < T{-}1$: pad by repeating $\hat{\mathbf{g}}_1$ \\
\quad 2.\ Forward pass (same as Training 1.a--e) $\rightarrow$ $\hat{\mathbf{g}}_t$ \\
\quad 3.\ Append $\hat{\mathbf{g}}_t$ to $\mathcal{C}$ \\
\end{tabular}
\caption{Training and inference procedure of EyeTAG.}
\label{alg:eyetag}
\end{algorithm}

To validate this design choice, we compare three strategies for constructing the gaze history while keeping all the other components fixed: (a) \emph{teacher forcing only} ($\alpha=1$ throughout), (b) \emph{fully autoregressive} ($\alpha=0$ throughout), and (c) our \emph{scheduled sampling} ($\alpha$ annealed from 1.0 to 0.0). We run this ablation with the normal gaze prior (w/ $\hat{\mathbf{G}}$) at $T{=}32$, the best-performing window length under the normal representation (see Sec.~\ref{sec:supp-tsweep-abs}), so that we isolate the effect of the sampling schedule from the gaze-history representation.

\begin{table}[t]
\centering
\footnotesize
\setlength{\tabcolsep}{4pt}
\resizebox{\linewidth}{!}{
\begin{tabular}{clcccc}
\toprule
& \textbf{Gaze History Schedule} & \textbf{All} & \textbf{Semi-Front} & \textbf{Front} & \textbf{Mean} \\
\midrule
(a) & Teacher forcing only ($\alpha=1$) & 39.53 $\pm$ 32.45 & 38.98 $\pm$ 31.81 & 31.64 $\pm$ 22.37 & 36.71 $\pm$ 28.87 \\
(b) & Fully autoregressive ($\alpha=0$) & \underline{10.13 $\pm$ 7.88} & \underline{9.96 $\pm$ 7.93} & \underline{8.81 $\pm$ 7.54} & \underline{9.63 $\pm$ 7.78} \\
\midrule
(\textbf{Ours}) & Scheduled sampling ($\alpha:1\!\rightarrow\!0$) & \textbf{9.63 $\pm$ 7.94} & \textbf{9.46 $\pm$ 7.38} & \textbf{8.22 $\pm$ 7.07} & \textbf{9.10 $\pm$ 7.46} \\
\bottomrule
\end{tabular}
}
\caption{Effect of the scheduled sampling strategy used to construct the gaze history during training, evaluated on Gaze360~\cite{kellnhofer2019gaze360} with the normal gaze prior (\textbf{w}/ $\hat{\mathbf{G}}$) at $T{=}32$. MAE (\textdegree); lower is better. The \textbf{best} and \underline{second-best} results in each column are marked in bold and underline, respectively.}
\label{tab:scheduled_sampling}
\end{table}

As reported in \cref{tab:scheduled_sampling}, teacher forcing only (a) is much worse at test time (36.71\textdegree{} Mean), because the model never sees its own prediction errors during training. This train--test mismatch makes the autoregressive errors accumulate quickly at inference. Fully autoregressive training (b) is much better than (a), but it is still worse than our scheduled sampling (c) by 0.53\textdegree{} in mean, because learning from noisy predictions in the early epochs makes the optimization harder. Scheduled sampling achieves the lowest error across all subsets, indicating that a gradual transition from teacher forcing to autoregressive histories is more effective than using either strategy alone

\section{Effect of Window Length with Normal Gaze Prior}
\label{sec:supp-tsweep-abs}

In the main paper (Sec.~5.3), we sweep the temporal window length $T$ under our proposed kinematic gaze prior (\textbf{w}/ $\Delta\hat{\mathbf{G}}$). Here we repeat the same sweep with the normal gaze prior (\textbf{w}/ $\hat{\mathbf{G}}$), which serves as a controlled baseline that separates the role of temporal context from the gaze-history representation. All experiments use the same architecture as our model (a VGGFace2-pretrained ResNet-50 face encoder, a shared ImageNet-pretrained ResNet-18 eye encoder, unidirectional cross-attention fusion, and a causal Transformer decoder), and we vary only $T$.

\begin{table}[t]
\centering
\footnotesize
\setlength{\tabcolsep}{4pt}
\begin{tabular}{ccccccc}
\toprule
\textbf{$T$} & \textbf{All} & \textbf{Semi-Front} & \textbf{Front} & \textbf{Mean} & \textbf{Time (ms)} & \textbf{GFLOPs} \\
\midrule
4  & 10.96 $\pm$ 8.28 & 10.80 $\pm$ 7.74 & 9.58 $\pm$ 7.42 & 10.45 $\pm$ 7.81 & 9.87 & 10.13 \\
8  & 10.17 $\pm$ 7.81 & 10.01 $\pm$ 7.28 & 9.23 $\pm$ 7.06 & 9.80 $\pm$ 7.38 & 9.97 & 20.26 \\
16 & 10.26 $\pm$ 7.98 & 10.08 $\pm$ 7.42 & 9.15 $\pm$ 7.29 & 9.83 $\pm$ 7.56 & 10.23 & 40.53 \\
24 & 9.96 $\pm$ 7.97  & 9.80 $\pm$ 7.36  & 8.65 $\pm$ 7.36 & 9.47 $\pm$ 7.56 & 11.45 & 60.79 \\
\textbf{32}$^{\ast}$ & \textbf{9.63 $\pm$ 7.94} & \textbf{9.46 $\pm$ 7.38} & \textbf{8.22 $\pm$ 7.07} & \textbf{9.10 $\pm$ 7.46} & 13.95 & 81.05 \\
48 & \underline{9.94 $\pm$ 7.94}  & \underline{9.65 $\pm$ 7.47}  & \underline{8.61 $\pm$ 7.65} & \underline{9.40 $\pm$ 7.69} & 19.98 & 121.57 \\
64 & 10.07 $\pm$ 8.16 & 9.91 $\pm$ 7.64  & 8.95 $\pm$ 7.61 & 9.64 $\pm$ 7.80 & 25.28 & 162.09 \\
\bottomrule
\end{tabular}
\caption{Effect of temporal window length $T$ on Gaze360~\cite{kellnhofer2019gaze360} under the normal gaze prior ($\textbf{w/}~\hat{\mathbf{G}}$). MAE (\textdegree); lower is better. The \textbf{best} and \underline{second-best} results in each column are marked in bold and underline, respectively.}
\label{tab:supp_tsweep_abs}
\end{table}

As reported in \cref{tab:supp_tsweep_abs}, the MAE decreases steadily from $T{=}4$ to $T{=}32$, which confirms that a longer temporal context provides useful information for gaze estimation. However, the accuracy starts to plateau and even degrade slightly beyond $T{=}32$. For example, the Semi-Front error rises from 9.46\textdegree{} at $T{=}32$ to 9.65\textdegree{} at $T{=}48$ and 9.91\textdegree{} at $T{=}64$. This saturation comes from the scale of the normal representation. As $T$ grows, the range of absolute pitch--yaw values expands together with the subject-specific bias, so it becomes harder for the model to extract consistent motion patterns from longer sequences. This motivates the kinematic gaze prior $\Delta\hat{\mathbf{G}}$ in the main paper (Sec.~5.2). The differencing removes subject-dependent gaze offsets, so the accuracy keeps improving at larger $T$.

\section{Robustness to Incomplete Gaze History}
\label{sec:supp-cold-start}

EyeTAG predicts the current gaze in an autoregressive manner, so the kinematic gaze history $\Delta\hat{\mathbf{G}}$ is only partially available during the first frames of a clip. As described in the main paper, for the initial $T-1$ frames, the differential gaze history is zero-padded, because a full window of past predictions does not yet exist. We therefore investigate how the model behaves in this cold-start setting, where the gaze history is incomplete.

For this analysis, we evaluate only the early frames of each clip, that is, the positions $0 \le t < T$ with $T=48$, over all 68 Gaze360 test clips~\cite{kellnhofer2019gaze360}. In this setting, the autoregressive history is only partially populated and is padded with zeros, so the kinematic prior is weakest here. As reported in \cref{tab:cold_start}, EyeTAG still achieves the lowest error (Mean $8.11$\textdegree), and even its absolute-prior variant, EyeTAG ($\textbf{w/}~\hat{\mathbf{G}}$), is second. The kinematic prior keeps its advantage over the Normal prior ($8.11$ vs $8.48$) and over all baselines even when the history is incomplete. This shows that the kinematic prior degrades gracefully and does not depend on a fully populated gaze history to remain effective.

\begin{table}[t]
\centering
\footnotesize
\setlength{\tabcolsep}{4pt}
\begin{tabular}{clcccc}
\toprule
\textbf{Type} & \textbf{Method} & \textbf{All} & \textbf{Semi-Front} & \textbf{Front} & \textbf{Mean} \\
\midrule
\multirow{3}{*}{SF} & L2CS-Net~\cite{abdelrahman2023l2cs} & 10.07 $\pm$ 7.66 & 9.85 $\pm$ 6.40 & 9.90 $\pm$ 6.37 & 9.94 $\pm$ 6.81 \\
& GazeTR-Hybrid~\cite{cheng2022gaze} & 9.42 $\pm$ 7.27 & 9.26 $\pm$ 6.50 & 9.46 $\pm$ 6.31 & 9.38 $\pm$ 6.69 \\
& CrossGaze~\cite{catruna2024crossgaze} & 9.22 $\pm$ 6.84 & 9.09 $\pm$ 6.05 & 8.55 $\pm$ 5.90 & 8.95 $\pm$ 6.26 \\
\midrule
\multirow{3}{*}{MF} & Gaze360~\cite{kellnhofer2019gaze360} & 9.70 $\pm$ 7.02 & 9.47 $\pm$ 6.00 & 9.41 $\pm$ 5.40 & 9.53 $\pm$ 6.14 \\
& STAGE~\cite{jindal2024spatio} & 8.87 $\pm$ 5.84 & 8.87 $\pm$ 5.84 & 8.69 $\pm$ 5.77 & 8.81 $\pm$ 5.82 \\
& EyeTAG ($\textbf{w/}~\hat{\mathbf{G}}$) & \underline{8.80 $\pm$ 6.41} & \underline{8.68 $\pm$ 6.08} & \underline{7.95 $\pm$ 6.22} & \underline{8.48 $\pm$ 6.23} \\
& \textbf{EyeTAG (Ours)} & \textbf{8.35 $\pm$ 5.92} & \textbf{8.27 $\pm$ 5.84} & \textbf{7.69 $\pm$ 5.69} & \textbf{8.11 $\pm$ 5.82} \\
\bottomrule
\end{tabular}
\caption{Robustness to an incomplete gaze history on Gaze360~\cite{kellnhofer2019gaze360}. We evaluate only the early frames of each test clip ($0 \le t < T$, $T=48$), where the autoregressive gaze history is only partially available and is zero-padded. MAE (\textdegree); lower is better. EyeTAG ($\textbf{w/}~\hat{\mathbf{G}}$) uses the normal gaze prior $\hat{\mathbf{G}}$, and EyeTAG (Ours) uses the kinematic prior $\Delta\hat{\mathbf{G}}$. The \textbf{best} and \underline{second-best} results in each column are marked in bold and underline, respectively.}
\label{tab:cold_start}
\end{table}

\end{document}